\documentclass{bmvc2k}

\usepackage{shortcuts}         

\newcommand{\outline}[1]{}
\usepackage{multirow}
\usepackage{graphicx}
\usepackage{float}
\usepackage{booktabs}
\usepackage{amsfonts}
\usepackage{amsmath}
\usepackage{wrapfig,lipsum}

\title{AegisRF: Adversarial Perturbations Guided with Sensitivity for Protecting Intellectual Property of Neural Radiance Fields}

\addauthor{Woo Jae Kim}{wkim97@kaist.ac.kr}{1}
\addauthor{Kyu Beom Han}{qbhan@kaist.ac.kr}{1}
\addauthor{Yoonki Cho}{yoonki@kaist.ac.kr}{2}
\addauthor{Youngju Na}{yjna2907@kaist.ac.kr}{3}
\addauthor{Junsik Jung}{junsik.jung@kaist.ac.kr}{4}
\addauthor{Sooel Son}{sl.son@kaist.ac.kr}{5}
\addauthor{Sung-eui Yoon}{sungeui@kaist.edu}{6}

\addinstitution{
 School of Computing,\\
 Korea Advanced Institute of Science and Technology (KAIST),\\
 Daejeon, Korea
}

\def\eg{\emph{e.g}\bmvaOneDot}

\def\ie{\emph{i.e}\bmvaOneDot}
\def\etal{\emph{et al}\bmvaOneDot}

\runninghead{Kim \etal}{AegisRF -- Adversarial Perturbations for NeRF IP Protection}

\begin{document}

\maketitle

\begin{abstract}
As Neural Radiance Fields (NeRFs) have emerged as a powerful tool for 3D scene representation and novel view synthesis, protecting their intellectual property (IP) from unauthorized use is becoming increasingly crucial. In this work, we aim to protect the IP of NeRFs by injecting adversarial perturbations that disrupt their unauthorized applications. However, perturbing the 3D geometry of NeRFs can easily deform the underlying scene structure and thus substantially degrade the rendering quality, which has led existing attempts to avoid geometric perturbations or restrict them to explicit spaces like meshes. To overcome this limitation, we introduce a \textit{learnable sensitivity} to quantify the spatially varying impact of geometric perturbations on rendering quality. Building upon this, we propose AegisRF, a novel framework that consists of a Perturbation Field, which injects adversarial perturbations into the pre-rendering outputs (color and volume density) of NeRF models to fool an unauthorized downstream target model, and a Sensitivity Field, which learns the sensitivity to adaptively constrain geometric perturbations, preserving rendering quality while disrupting unauthorized use. Our experimental evaluations demonstrate the generalized applicability of AegisRF across diverse downstream tasks and modalities, including multi-view image classification and voxel-based 3D localization, while maintaining high visual fidelity. 
Codes are available at \url{https://github.com/wkim97/AegisRF}.
\end{abstract}
\section{Introduction}
\label{sec:intro}
Neural Radiance Fields (NeRFs)~\cite{nerf} have emerged as a powerful paradigm for novel view synthesis and 3D scene representation, finding applications in AR/VR~\cite{vr-nerf, instant-3d}, autonomous driving~\cite{neurad, nerf-autonomous-survey}, and the metaverse~\cite{ue4-nerf, museum}.
Their implicit representation, which yields pre-rendering outputs (\ie, color, volume density) for any queried 3D point and viewing direction, has enabled scene representation in diverse data modalities such as images, voxels, and point clouds, thus driving advancements in diverse downstream 3D perception tasks~\cite{nerf-rpn, nerf-det, sa3d, nesf, instance-nerf}.
However, their broad applicability entails a new challenge: protecting their intellectual property (IP) from unauthorized use.
Given substantial resources required to construct radiance fields~\cite{pixelnerf, coponerf, instantsplat, ingp, 3dgs, share} and their value as a versatile 3D data source~\cite{ue4-nerf, vr-nerf, neurad}, their unauthorized use in these downstream tasks can lead to significant losses in resources and revenues, making IP protection an urgent concern~\cite{copyrnerf, waterf, geometrycloak, geometrysticker}.

Inspired by the recent success of adversarial perturbations for IP and privacy protection in text~\cite{memguard, attriguard, aaad}, audio~\cite{voicecamouflage, voiceblock}, and image~\cite{photoguard, glaze, advdm, advface, clip2protect, fawkes, advpaint} domains, we propose to tackle this challenge by obstructing downstream models that attempt to exploit NeRFs without proper authorization.
We achieve this by introducing carefully crafted adversarial perturbations into the NeRF's pre-rendering outputs (color and density) designed to undermine an unauthorized target downstream model while preserving the visual quality.
By injecting these perturbations at inference time, our approach can generate adversarial examples across diverse data modalities derived from NeRFs, thus offering a generalized protection framework applicable to a wide range of downstream tasks.

\begin{figure}[t]
    \centering
    \includegraphics[width=\columnwidth]{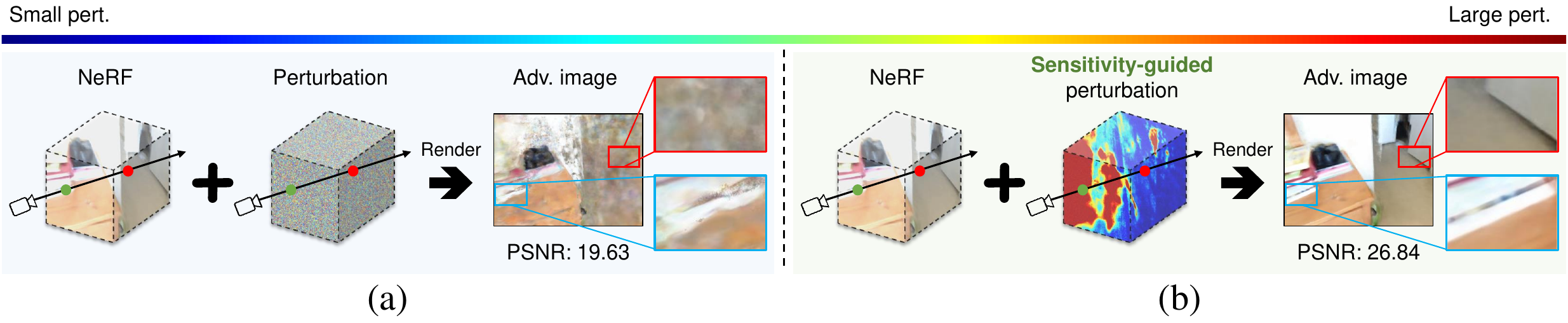}
    \vspace{-20pt}
    \caption{
    (a) Geometric perturbations without consideration of their varying impact on rendering quality lead to significant degradation in rendering quality.
    (b) Our novel approach mitigates this by measuring the sensitivity of rendering quality to geometric perturbations and adaptively constraining their magnitudes.
    For example, perturbations are restricted on empty spaces (\textcolor[HTML]{FF0000}{red point}), where disruptions can cause introduction of new artifacts, while larger perturbations are applied on more complex regions (\textcolor[HTML]{70AD47}{green point}), where such disruptions can be better masked by the existing structural complexity.
    }
    \vspace{-10pt}
    \label{fig:motivation}
\end{figure}

However, perturbing the 3D geometry of NeRFs is inherently challenging, as it can easily deform the underlying scene structure and substantially degrade rendering quality~\cite{tt3d, copyrnerf, stylizednerf}.
We argue that this degradation primarily occurs because the visual impact of a geometric perturbation is not uniform across the 3D space; rather, it heavily depends on the specific 3D location and the local structural details~\cite{legge, thicket, vq-assessment}.
For example, as shown in Fig.~\ref{fig:motivation}(a), applying perturbations of random magnitudes across the 3D space without considering their spatially varying impact can indiscriminately alter geometric structures and thus significantly degrade the rendering quality.
While there have been recent attempts to inject adversarial perturbations into NeRFs~\cite{tt3d, nerfail}, they also overlook this insight, either imposing no explicit constraints on perturbations~\cite{tt3d} or applying predetermined, fixed constraints across 3D space~\cite{nerfail}.
Consequently, to prevent visual distortions, these methods resort to either avoiding geometric perturbations altogether~\cite{nerfail} or restricting them to explicit forms like meshes~\cite{tt3d}, thereby limiting their applicability across the diverse 3D downstream tasks that leverage NeRF's pre-rendering outputs.

To overcome this limitation when perturbing the geometry of NeRFs, we introduce a \textit{learnable sensitivity} measure for quantifying the varying degree of impact that geometric perturbations across 3D space have on rendering quality.
As shown in Fig.~\ref{fig:motivation}(b), this allows us to impose tighter constraints on geometric perturbations in regions where changes would be highly perceptible, while permitting more perturbations in areas where they cause less perceptual degradation.
This sensitivity-guided strategy aims to maximize the disruptive effect on unauthorized tasks while minimizing degradation in rendering quality.

Based on this insight, we propose \sys, a novel framework for protecting the IP of NeRF models by injecting sensitivity-guided perturbations into their pre-rendering outputs at inference time.
AegisRF consists of two key components: (1) Perturbation Field that generates appearance (color) and geometry (density) perturbations designed to fool an unauthorized downstream target model, and (2) Sensitivity Field that quantifies the sensitivity of the rendering quality to geometric perturbations across 3D space and constrains these perturbations adaptively.
Our experimental evaluations demonstrate that \sys offers robust protection across diverse downstream tasks, including multi-view image classification~\cite{vit} and voxel-based 3D localization~\cite{nerf-rpn}, while preserving the rendering quality (Sec.~\ref{ssec:main-results}).
Additionally, through extensive analysis on the Sensitivity Field (Sec.~\ref{ssec:analysis}), we verify that guiding the perturbations with the sensitivity is crucial to maintain high rendering quality.

In summary, our contributions are as follow:

\hangindent=1em \noindent\hspace{0.5em}\textbullet$\,\,$We introduce learnable sensitivity for measuring the perceptual impact of geometric perturbations on rendering quality, which enables adaptive constraints for robust NeRF IP protection while maintaining high visual fidelity.

\hangindent=1em \noindent\hspace{0.5em}\textbullet$\,\,$We present \sys, a novel framework for NeRF IP protection that operates at inference time by perturbing both the appearance and geometry pre-rendering outputs of NeRFs, thus protecting their IP from a diverse range of downstream tasks and modalities.

\hangindent=1em \noindent\hspace{0.5em}\textbullet$\,\,$Through empirical evaluations, we verify the effectiveness of \sys at substantially undermining the performance of various downstream applications with different modalities while preserving the rendering quality.

\section{Related Works}
\vspace{0.5ex}\noindent
\textbf{Neural Radiance Fields.}
Neural radiance fields (NeRFs)~\cite{nerf} and other MLP-based radiance fields~\cite{depth-nerf, tensorf, kilonerf, donerf, ingp, coponerf}, implicitly represent 3D scenes as continuous signals. These methods have achieved photorealistic novel view synthesis and have become a powerful 3D representation~\cite{depth-nerf, tensorf, kilonerf, donerf, ingp, uforecon}.
Their implicit nature allows transformation of their pre-rendering outputs into various data types (\eg, images, voxels)~\cite{sa3d, perfception, nerf-rpn, nerf-det}, fueling diverse downstream perception tasks such as classification~\cite{vit, nerfail, perfception}, segmentation~\cite{nesf, instance-nerf, sa3d, nerf-sos, gp-nerf}, and localization~\cite{nerf-rpn, nerf-det}.
As commercial NeRF applications also grow~\cite{nerfstudio}, protecting their IP from unauthorized downstream use becomes critical.
This work safeguards IP of NeRFs across various downstream tasks and modalities by applying adversarial perturbations to pre-rendering outputs while ensuring visual integrity.

\vspace{0.5ex}\noindent
\textbf{IP Protection via Model Deception.}
Protecting data IP or privacy by subverting downstream applications with imperceptible input perturbations has become a popular strategy.
This includes deceiving facial recognition~\cite{advface, clip2protect, fawkes}, preventing unauthorized manipulation or style extraction from generative models~\cite{photoguard, glaze, advdm, advpaint}, and protecting text or audio data~\cite{memguard, attriguard, aaad, voicecamouflage, voiceblock}.
We extend this paradigm of IP protection through model deception to NeRFs~\cite{nerf}, aiming to thwart unauthorized use by designing adversarial attacks targeting their downstream applications.

\vspace{0.5ex}\noindent
\textbf{3D Adversarial Attack.}
Since Athalye \etal~\cite{eot} first proposed 3D adversarial examples, many attacks have targeted diverse data forms like point clouds~\cite{3dadv, shape-3dadv, advpc, isometry, lidar-seg, pc-seg, pc-det, lidar-det}, meshes~\cite{mc-mesh-attack, renderer-attack, mesh-attack, saga}, and voxel-grids~\cite{3d-shape-classifier}. 
Recently, the rise of radiance fields~\cite{nerf, depth-nerf, ingp} spurred NeRF-specific attacks; TT3D~\cite{tt3d} adversarially fine-tunes radiance fields~\cite{ingp}, and NeRFail~\cite{nerfail} perturbs the color attributes of NeRFs from transformed 2D pixels.
However, these works overlook the sensitivity of geometry to perturbations, thus avoiding~\cite{nerfail} or limiting geometric perturbations to vertex coordinates on the explicit mesh space~\cite{tt3d} to avoid significant deformations.
In contrast, our sensitivity-guided approach provides a spatially-aware understanding of how geometric perturbations affect rendering quality, thereby enabling direct perturbations on NeRF's implicit geometry for versatile IP protection across different downstream tasks and modalities while preserving rendering quality.

\begin{figure*}
    \centering
    \includegraphics[width=0.95\linewidth]{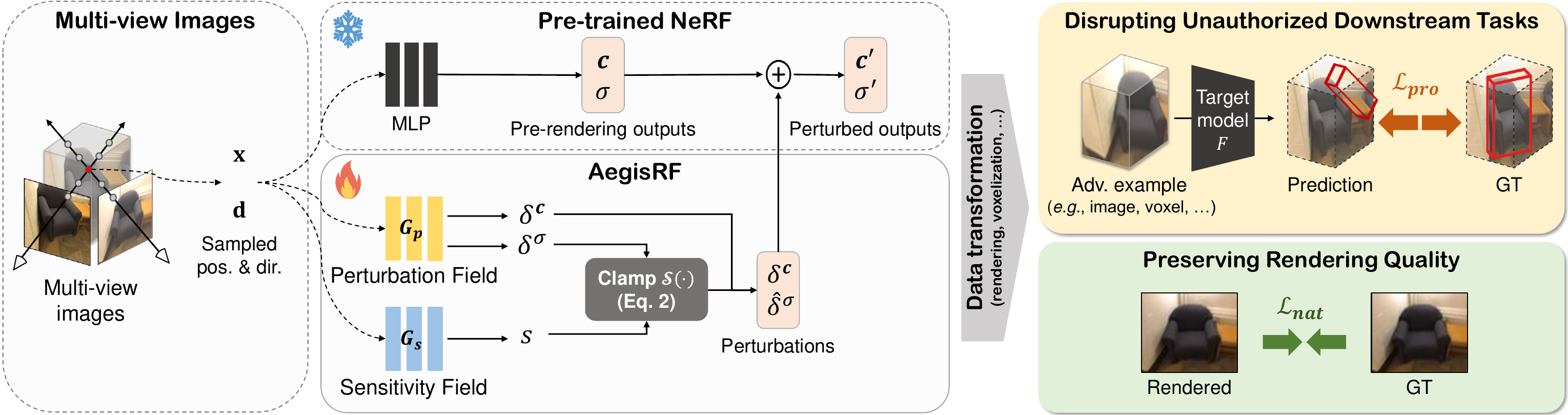}
    \vspace{-5pt}
    \caption{
    Overview of \sys.
    For a 3D point $(\mathbf{x}, \mathbf{d})$, the Perturbation Field creates appearance ($\delta^\mathbf{c}$) and geometry ($\delta^\sigma$) perturbations, while the Sensitivity Field predicts sensitivity ($s$) to adaptively constrain geometry perturbations ($\hat{\delta}^\sigma$).
    These perturb NeRF outputs ($\mathbf{c}$, $\sigma$) into perturbed versions ($\mathbf{c}'$, $\sigma'$), forming adversarial examples in various data forms, aiming to disrupt target unauthorized downstream task ($\mathcal{L}_{\text{pro}}$) while preserving rendering quality ($\mathcal{L}_{\text{nat}}$).
    }
    \vspace{-10pt}
    \label{fig:overview}
\end{figure*}

\section{\sys}
\subsection{Preliminaries}
\label{sec:prelim}
\vspace{0.5ex}\noindent
\textbf{Neural Radiance Fields (NeRF).}
NeRF~\cite{nerf} models a 3D scene using an MLP that maps a 3D coordinate $\mathbf{x} \in \mathbb{R}^3$ and view direction $\mathbf{d} \in \mathbb{R}^3$ to pre-rendering outputs: RGB color $\mathbf{c}$ and volume density $\sigma$, \ie, $\text{MLP}(\mathbf{x}, \mathbf{d}) = (\mathbf{c}, \sigma)$.
Images are rendered by volumetrically integrating these outputs along rays $\mathbf{r}(t) = \mathbf{o} + t\mathbf{d}$ emitted from the camera center $\mathbf{o}$.
For $N$ points sampled along each ray, the pixel color $I(\mathbf{r})$ is computed as:
\begin{equation}
    I(\mathbf{r}) = \sum_{i=1}^N T_i (1 - \exp(-\sigma_i \Delta t_i)) \mathbf{c}_i,
    \label{eq:render_nerf}
\end{equation}
where $T_i = \exp\left(-\sum_{j=1}^{i-1} \sigma_j \Delta t_j\right)$ is accumulated transmittance, $\Delta t_i = t_{i+1} - t_i$ is the distance between consecutive sampled points, and $\mathbf{c}_i$, $\sigma_i$ are the color and density at the $i$-th sample.

\subsection{Perturbation and Sensitivity Fields}
\label{ssec:pert-sens-field}
We tackle the research question of protecting IP in NeRFs from adversaries who use these radiance fields without proper authorization.
To achieve this, we propose \sys (Fig.~\ref{fig:overview}), a framework that undermines the adversary's downstream task by injecting imperceptible perturbations into the pre-rendering outputs of NeRFs, or the appearance (\ie, color $\mathbf{c}$) and the geometry (\ie, density $\sigma$).
\sys consists of two components: (1) Perturbation Field, which models adversarial perturbations designed to undermine the target unauthorized downstream model, and (2) Sensitivity Field, which constrains the geometric perturbations based on the degree of impact they have on the rendering quality degradation.

\vspace{0.5ex}\noindent
\textbf{Perturbation Field.}
To protect NeRFs from unauthorized usage, we introduce the \emph{Perturbation Field} (Fig.~\ref{fig:overview}) that perturbs their color $\mathbf{c}$ and density $\sigma$ outputs by injecting appearance perturbation $\delta^\mathbf{c}$ and geometry perturbation $\delta^\sigma$ such that $\mathbf{c}' = \mathbf{c} + \delta^\mathbf{c}$ and $\sigma' = \sigma + \delta^\sigma$.
We model the Perturbation Field via a set of multilayer perceptron (MLP) networks $G_p$ to learn $\mathbf{\delta}^\mathbf{c}$ and $\delta^\sigma$ given a point with position $\mathbf{x}$ and viewing direction $\mathbf{d}$ from a camera such that $G_p : (\mathbf{x}, \mathbf{d}) \rightarrow (\delta^\mathbf{c}, \delta^\sigma)$.
These perturbations are applied to the pre-rendering outputs of a frozen pre-trained NeRF at inference time.
This allows our perturbations to be easily activated or deactivated depending on the deployment context, enabling seamless switching between protected and unprotected inference modes.
More details on the architecture of $G_p$ and sampling methods are provided in the appendix (Sec.~\ref{sec:additional-details}).

The implicit design of our Perturbation Field allows us to craft perturbations for any queried 3D point and view, making it highly compatible with a wide range of MLP-based radiance fields.
These perturbations injected directly into the pre-rendering outputs of NeRFs can also be transformed into adversarial examples of diverse modalities, such as voxel grids~\cite{nerf-rpn, nesf, nerf-det} or images~\cite{perfception, sa3d}.
Such versatility allows application of \sys across diverse unauthorized downstream models (Sec.~\ref{ssec:main-results}), providing generalized IP protection.

\vspace{0.5ex}\noindent
\textbf{Sensitivity Field.}
Unlike appearance, even small disruptions on the 3D geometry can lead to significant shape deformations and thus degrade the rendering quality~\cite{tt3d, copyrnerf, stylizednerf}.
To this end, we design the \textit{Sensitivity Field} that measures the sensitivity of rendering quality to geometric perturbations across the 3D space and impose constraints accordingly.
We approximate the Sensitivity Field via a set of MLPs $G_s$ that takes the position $\mathbf{x}$ of a point and outputs a sigmoid-normalized scalar sensitivity $s \in [0, 1]$ such that $G_s : \mathbf{x} \rightarrow s$ as shown in Fig.~\ref{fig:overview}.
Given the sensitivity $s$, we constrain the magnitudes of the geometry perturbation, thus applying a tighter constraint on regions that are sensitive to rendering quality degradation while allowing larger perturbations on regions with lower sensitivity scores.

While a common strategy for such constraints would be clipping the perturbation as commonly done in traditional norm-bounded attacks~\cite{fgsm, pgd, ada, fsr}, this could prevent the gradient flow and interfere with the training of Perturbation and Sensitivity Fields.
Thus, we use the soft clamping strategy $\mathcal{S}(\cdot, \cdot)$ to obtain constrained geometric perturbation $\hat{\delta}^\sigma$ while ensuring their proper training as follows:
\begin{align}
    \hat{\delta}^\sigma &= \mathcal{S}(\delta^\sigma, s) = ((1-s) \cdot \bar{\sigma}) \cdot \tanh\left(\frac{\delta^\sigma}{(1-s) \cdot \bar{\sigma}}\right),
    \label{eq:soft-clamp}
\end{align}
where $\bar{\sigma}$ is the mean density value for all points uniformly sampled from a 3D grid that covers the entire scene volume.
With this soft clamping operation, adversarial color $\mathbf{c}'$ and density $\sigma'$ can now be written as $\mathbf{c}' = \mathbf{c} + \mathbf{\delta^\mathbf{c}}$ and $\sigma' = \sigma + \hat{\delta}^\sigma$.

Traditional norm-bounded image attacks~\cite{fgsm, pgd} use fixed perturbation magnitudes, which overlook the varying perceptibility of geometric perturbations in 3D space.
In contrast, our method adapts constraints based on the 3D location, applying tighter limits in sensitive areas and allowing larger perturbations in less sensitive regions to ensure imperceptibility compared to fixed constraints (Sec.~\ref{ssec:analysis}).

\subsection{Training \sys}
\label{ssec:objectives}
We train \sys using two objectives: (1) protection loss, which guides the generation of strong perturbations that can undermine the target unauthorized downstream task, and (2) naturalness loss, which aims to preserve rendering quality in the presence of perturbations.

\vspace{0.5ex}\noindent
\textbf{Protection loss $\mathcal{L}_{\text{pro}}$.}
The primary objective of the protection loss is to guide the Perturbation Field $G_p$ to produce perturbations that effectively undermine a target unauthorized downstream model $F$.
To achieve this, we first generate an adversarial example $x'$, which serves as the input to $F$.
Thanks to the versatile characteristic of the pre-rendering outputs of NeRF, the adversarial example $x'$ can also be generated in different modalities from the perturbed pre-rendering outputs $(\mathbf{c}', \sigma')$ depending on the downstream model.
For instance, if $F$ is an image-based model~\cite{vit, nerfail}, $x'$ is an RGB image produced by volumetric rendering (Eq.~\ref{eq:render_nerf}) of $(\mathbf{c}', \sigma')$, and if $F$ is a voxel-based model~\cite{nerf-rpn}, $x'$ is a 3D voxel grid constructed via uniform sampling and aggregation~\cite{nerf-rpn, nerf-det}, where each point in grid captures $(\mathbf{c}', \sigma')$.
This versatility in forming $x'$ allows our approach to target a wide range of downstream tasks that utilize various data modalities derivable from NeRFs.

The protection loss $\mathcal{L}_{\text{pro}}$ is defined to maximize the training loss $\mathcal{L}_F$ of the target model $F$ (\eg, cross-entropy for classification) as follows:
\begin{equation}
    \mathcal{L}_{\text{pro}} = - \mathcal{L}_F( F(x'), y_{gt} ),
    \label{eq:attack-loss}
\end{equation}
where $F(x')$ is the prediction from the target downstream model, and $y_{gt}$ is the corresponding ground-truth label (\eg, a class label for classification).
In this way, the Perturbation Field learns to generate perturbations $(\delta^\mathbf{c}, \delta^\sigma)$, which are then used to craft perturbed outputs $(\mathbf{c}', \sigma')$ and ultimately the adversarial example $x'$ that is highly effective at degrading the performance of unauthorized models.

\vspace{0.5ex}\noindent
\textbf{Naturalness loss $\mathcal{L}_{\text{nat}}$.}
To maintain rendering quality against perturbations, our naturalness loss $\mathcal{L}_{\text{nat}}$ measures the photometric $L_2$ difference~\cite{nerf} between a rendering output $I'_n$ derived from perturbed color $\mathbf{c}'$ and density $\sigma'$ (Eq.~\ref{eq:render_nerf}) and its corresponding ground truth $I_n^{gt}$.
We compute the loss over a set of camera rays $\mathcal{R}$ sampled from the training views:
\begin{equation}
    \mathcal{L}_{\text{nat}} = \sum_{\mathbf{r} \in \mathcal{R}} \parallel I'_n(\mathbf{r}) - I_n^{gt}(\mathbf{r}) \parallel_2^2.
    \label{eq:naturalness-loss}
\end{equation}
This term penalizes rendering quality degradation and supervises both the Perturbation Field $G_p$ and the Sensitivity Field $G_s$.
In this way, the Perturbation Field $G_p$ learns to refrain from generating perturbations $(\delta^\mathbf{c}, \delta^\sigma)$ that result in severe degradation in rendering quality.
This term also guides the Sensitivity Field $G_s$ to predict high sensitivity values on regions where geometric perturbations would significantly degrade rendering quality, and low sensitivity to regions where perturbations are perceptually tolerable, thus serving our purpose of crafting effective geometric perturbations while preserving the rendering quality.

\vspace{0.5ex}\noindent
\textbf{Model training.}
During training, the pre-trained NeRF remains frozen while our Perturbation and Sensitivity Fields are updated on the combination of the two losses:
\begin{equation}
    \mathcal{L} = \lambda_{\text{pro}} \cdot \mathcal{L}_{\text{pro}} + \lambda_{\text{nat}} \cdot \mathcal{L}_{\text{nat}},
\end{equation}
where $\lambda_{\text{pro}}$ and $\lambda_{\text{nat}}$ denote the coefficients of $\mathcal{L}_{\text{pro}}$ and $\mathcal{L}_{\text{nat}}$, respectively.
\section{Experiments}
\subsection{Experimental Setups}
\vspace{0.5ex}\noindent
\textbf{Dataset and evaluation metrics.}
We use $8$ scenes from the NeRF Synthetic dataset~\cite{nerf} for multi-view classification and $10$ scenes from the ScanNet~\cite{scannet} dataset for 3D localization.
We use the train/test view splits set by the target downstream methods.
We evaluate naturalness with PSNR, SSIM, and LPIPS~\cite{lpips} of rendered images~\cite{nerf, dvgo, tensorf}. 
We also assess disruptive efficacy on downstream task, \ie, the protective performance of \sys.
For multi-view classification~\cite{vit}, we measure prediction accuracy on images rendered on test views.
For 3D localization~\cite{nerf-rpn}, we measure Recall@K and AP@K of predicted bounding boxes.

\vspace{0.5ex}\noindent
\textbf{Target models and baseline methods.}
We evaluate our \sys on two MLP-based radiance fields and two representative downstream tasks: (1) multi-view classification (ViT-B/16~\cite{vit} on NeRF~\cite{nerf}) and (2) voxel-based 3D localization (NeRF-RPN~\cite{nerf-rpn} with Swin Transformer~\cite{swin} on depth-guided NeRF~\cite{depth-nerf}).
As a baseline, we consider TT3D~\cite{tt3d}, a method that adversarially fine-tunes parameters of a pre-trained radiance field.
However, while TT3D modifies the appearance parameters of the radiance field, it alters geometry on the explicit mesh derived from the radiance field.
Since our goal is to protect the NeRF model itself, which can form diverse data modalities, TT3D's focus on a single derived mesh is not directly suitable.
Thus, we adapt TT3D’s core idea of adversarially fine-tuning into Adv-FT. This approach fine-tunes the parameters of the pre-trained NeRF to generate both the adversarial color and density representations via the protection loss (Eq.~\ref{eq:attack-loss}) and the naturalness loss (Eq.~\ref{eq:naturalness-loss}).
For multi-view classification, we also compare with NeRFail~\cite{nerfail} and NeRFail-S~\cite{nerfail}, which are specifically designed for multi-view adversarial attacks.

\vspace{0.5ex}\noindent
\textbf{Implementation details.}
For \sys, we set protection loss weight $\lambda_\text{pro} = 0.0003$, naturalness loss weight $\lambda_\text{nat} = 1$ for multi-view classification and $\lambda_\text{pro} = 1$, $\lambda_\text{nat} = 50$ for 3D localization.
For Adv-FT, we set $\lambda_\text{nat} = 1$ for multi-view classification and $\lambda_\text{nat} = 50$ for 3D localization, and use various values of $\lambda_\text{pro}$ for extensive comparisons over its performance spectrum.
For NeRFail~\cite{nerfail}, we used hyperparameters set by the authors and $\epsilon=32$.
Please refer to Sec.~\ref{sec:additional-details} of the supplementary for additional implementation details.

\begin{table*}[t]
    \centering
    \resizebox{0.9\linewidth}{!}{
    \begin{tabular}{c|ccc|cccc}
    \specialrule{2pt}{\aboverulesep}{\belowrulesep}
    \multicolumn{8}{c}{\textbf{Multi-view classification}} \\
    \specialrule{1pt}{\aboverulesep}{\belowrulesep}
    \multirow{2}{*}{Method} & \multicolumn{3}{c|}{Naturalness} & \multicolumn{4}{c}{Disruption efficacy} \\
    \cline{2-8}
    
    & PSNR $(\uparrow)$ & SSIM $(\uparrow)$ & LPIPS $(\downarrow)$ & \multicolumn{4}{c}{Acc. (\%)} \\
    \specialrule{1pt}{\aboverulesep}{\belowrulesep}
    
    No protection & 28.28 & 0.9205 & 0.0979 & \multicolumn{4}{c}{99.00} \\
    \midrule
    NeRFail-S & 22.42 & 0.8360 & 0.1508 & \multicolumn{4}{c}{2.00} \\
    NeRFail & 23.74 & 0.8751 & 0.1315 & \multicolumn{4}{c}{7.75} \\
    Adv-FT ($\lambda_\text{pro} = 0.0003$) & \underline{25.57} & \underline{0.8973} & \underline{0.1372} & \multicolumn{4}{c}{5.50} \\
    Adv-FT ($\lambda_\text{pro} = 0.003$) & 23.67 & 0.8722 & 0.1987 & \multicolumn{4}{c}{2.13} \\
    Adv-FT ($\lambda_\text{pro} = 0.03$) & 17.62 & 0.7442 & 0.5220 & \multicolumn{4}{c}{\textbf{0.13}} \\
    \sys(Ours) & \textbf{26.33} & \textbf{0.9042} & \textbf{0.1271} & \multicolumn{4}{c}{\underline{1.88}} \\
    \specialrule{2pt}{\aboverulesep}{\belowrulesep}

    \multicolumn{8}{c}{\textbf{3D localization}} \\
    \specialrule{1pt}{\aboverulesep}{\belowrulesep}
    \multirow{2}{*}{Method} & \multicolumn{3}{c|}{Naturalness} & \multicolumn{4}{c}{Disruption efficacy} \\
    \cline{2-8}
    
    & PSNR $(\uparrow)$ & SSIM $(\uparrow)$ & LPIPS $(\downarrow)$ & $\text{Recall}_{25}$ (\%) & $\text{Recall}_{50}$ (\%) & $\text{AP}_{25}$ (\%) & $\text{AP}_{50}$ (\%) \\
    \specialrule{1pt}{\aboverulesep}{\belowrulesep}
    
    No protection & 23.98 & 0.7884 & 0.3186 & 95.44 & 66.63 & 59.94 & 44.35 \\
    \midrule
    Adv-FT ($\lambda_\text{pro} = 0.1$) & \underline{22.37} & \underline{0.7249} & \underline{0.4289} & 66.58 & 15.86 & 14.93 & 2.52 \\
    Adv-FT ($\lambda_\text{pro} = 1$) & 18.44 & 0.6068 & 0.6013 & 50.82 & 6.88 & 3.83 & 0.22 \\
    Adv-FT ($\lambda_\text{pro} = 10$) & 15.10 & 0.5264 & 0.7147 & \textbf{43.43} & \textbf{2.25} & \textbf{1.60} & \textbf{0.01} \\
    \sys(Ours) & \textbf{23.97} & \textbf{0.7687} & \textbf{0.3565} & \underline{48.64} & \underline{4.77} & \underline{3.24} & \underline{0.20} \\
    
    \specialrule{2pt}{\aboverulesep}{\belowrulesep}
    
    \end{tabular}
    }
    \caption{
    Comparison of our \sys with existing methods on multi-view image classification and 3D localization.
    Best results are in \textbf{bold}, and second best results are \underline{underlined}.
    }
    \vspace{-10pt}
    \label{tab:main-table}
\end{table*}

\subsection{Naturalness and Protection Performance}
\label{ssec:main-results}
In Table~\ref{tab:main-table}, we report the naturalness and disruption efficacy of our \sys along with the compared methods.
Compared to NeRFail~\cite{nerfail}, \sys achieves both higher visual quality ($+2.59$ PSNR) and stronger protection ($-5.87\%$p accuracy) in multi-view classification.
This is thanks to our effective geometric perturbations, while NeRFail limits perturbations only to appearance components of NeRFs.
We can also observe that Adv-FT faces a large trade-off between the naturalness and disruption efficacy, achieving high disruption efficacy (\eg, $1.60\%$ AP$_{25}$ in 3D localization with $\lambda_\text{pro} = 10$) at a cost of significantly degraded rendering quality ($15.10$ PSNR).
Conversely, if Adv-FT prioritizes naturalness ($22.37$ PSNR with $\lambda_\text{pro} = 0.1$), this sacrifices its disruption efficacy ($14.93\%$ AP$_{25}$).
In contrast, our \sys achieves strong disruption efficacy ($3.24\%$ AP$_{25}$) with high visual quality ($23.97$ PSNR) thanks to the sensitivity guidance, which adaptively constrains geometric perturbations to preserve visual fidelity while effectively disrupting downstream models.

\begin{wraptable}{r}{0.35\linewidth}
    \centering
    \vspace{-10pt}
    \resizebox{\linewidth}{!}{%
    \begin{tabular}{c|ccc}
        \toprule
        \multirow{2}{*}{Surrogate Model} & \multicolumn{3}{c}{Target Accuracy (\%)} \\
        \cmidrule{2-4}
        & ViT-B & Swin-B & Mixer-B \\
        \midrule
        No Protection & 99.00 & 99.88 & 90.75 \\
        ViT-B         & 1.88  & 41.00 & 44.75 \\
        Swin-B        & 19.63 & 0.00  & 18.75 \\
        Mixer-B       & 7.25  & 6.00  & 0.50 \\
        \bottomrule
    \end{tabular}%
    }
    \vspace{-5pt}
    \caption{Transferability of \sys from a surrogate model to unknown target models in multi-view classification.}
    \label{tab:transferability-main}
    \vspace{-10pt}
\end{wraptable}

We further evaluate \sys in a more practical scenario by evaluating its transferability across different architectures in Table~\ref{tab:transferability-main}.
\sys demonstrates considerable transferability, reducing the multi-view classification accuracy of Swin-B to $6.00\%$ when trained to disrupt Mixer-B, highlighting the practical effectiveness of \sys in real-world IP protection scenarios.
In the supplementary, we report transferability on 3D localization (Sec.~\ref{sec:transferability}), robustness against common transformations such as JPEG compression or Gaussian noise (Sec.~\ref{sec:common-transformations}), and computational costs (Sec.~\ref{sec:addtional-analysis}).

\begin{figure*}[t]
    \centering
    \includegraphics[width=0.95\linewidth]{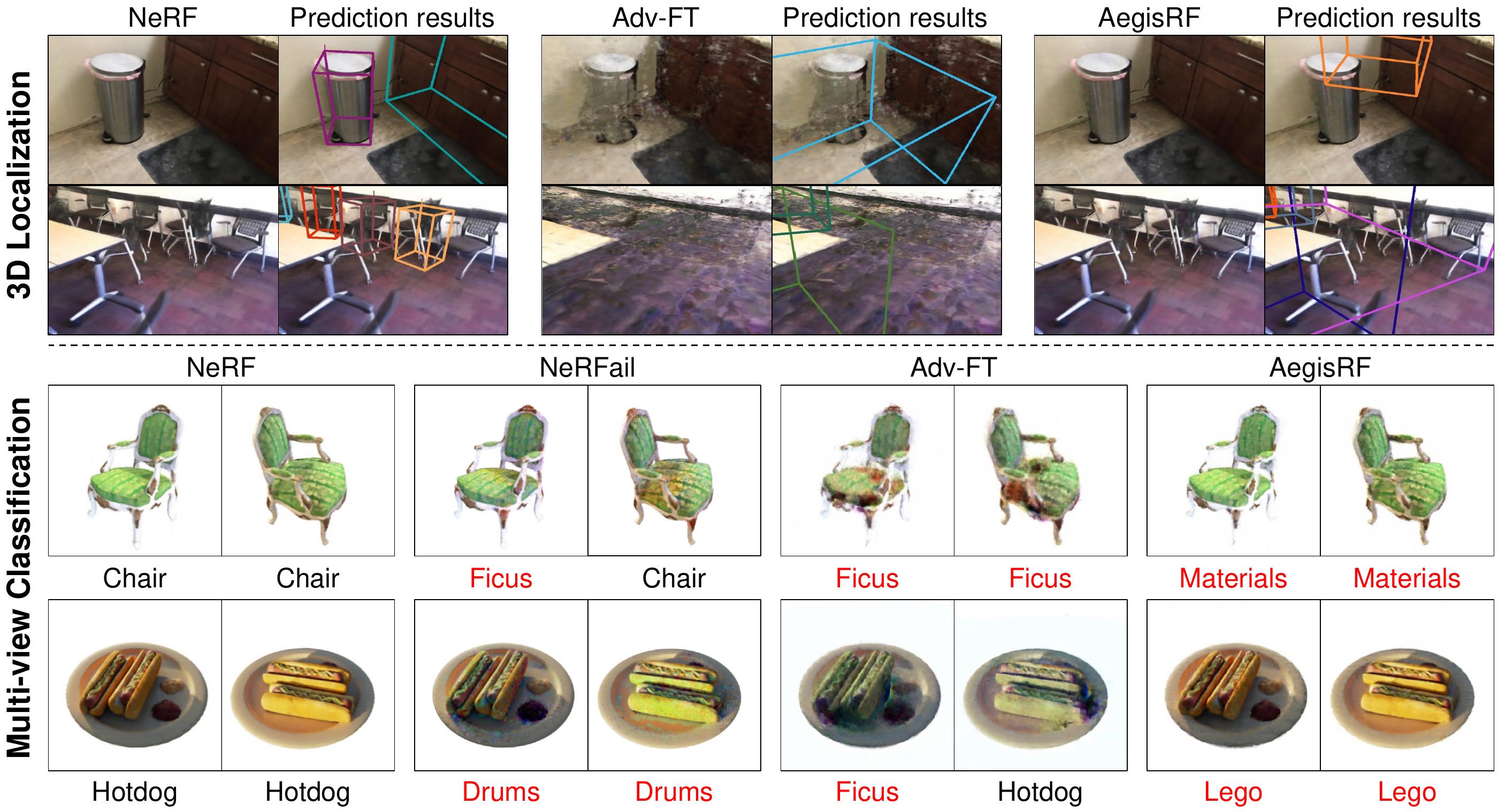}
    \vspace{-10pt}
    \caption{
    Visualizations of rendered images and model predictions on NeRF, NeRFail, Adv-FT ($\lambda_\text{pro} = 1$ for 3D localization, $\lambda_\text{pro} = 0.003$ for multi-view classification), and our \sys.
    Our \sys shows superior rendering quality compared to NeRFail and Adv-FT.
    }
    \vspace{-10pt}
    \label{fig:rendered-viz}
\end{figure*}

In Fig.~\ref{fig:rendered-viz}, we visualize images rendered from the original NeRF, existing methods, and our \sys along with their disruption efficacy.
While Adv-FT shows strong protective ability, it results in significantly distorted rendering qualities, often showing visible distortions in the geometry (\eg, removal of chairs in row 2).
In contrast, \sys can preserve the visual fidelity while undermining the predictions of the target models, thanks to the Perturbation and Sensitivity Fields, which balance adversarial strength with imperceptibility.

\subsection{Analysis on Sensitivity Field}
\label{ssec:analysis}
\begin{figure}[t]
    \centering
    \includegraphics[width=\linewidth]{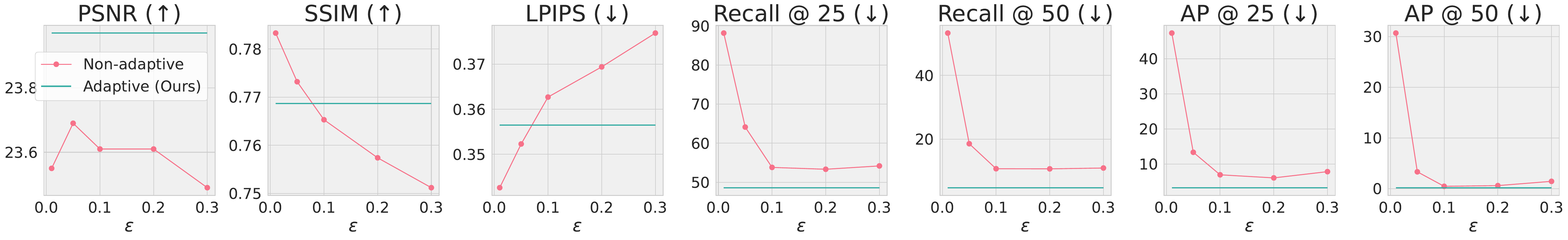}
    \vspace{-20pt}
    \caption{
    Comparison of naturalness (col. 1-3) and disruption efficacy (col. 4-7) of our sensitivity-guided adaptive approach with a fixed, non-adaptive perturbation bound $\epsilon$.
    Our strategy leads to the best balance between naturalness and protection performance.
    }
    \vspace{-5pt}
    \label{fig:fixed-eps}
\end{figure}

\vspace{0.5ex}\noindent
\textbf{Sensitivity vs. fixed bound.}
We compare Sensitivity Field with constraining geometric perturbations using a fixed, non-adaptive perturbation bound $\epsilon$ commonly used in traditional norm-bounded adversarial examples~\cite{fgsm, pgd}.
As shown in Fig.~\ref{fig:fixed-eps}, our sensitivity-guided adaptive approach outperforms all of the non-adaptive cases in terms of disruption efficacy.
While lower $\epsilon$ improves naturalness of non-adaptive approach, this comes at significantly degraded disruption efficacy.
In contrast, thanks to the sensitivity-guided approach that suppresses perturbations detrimental to rendering quality while perturbing less visually critical points, our method achieves a better balance between the naturalness and disruption efficacy.

\begin{wrapfigure}{r}{0.5\linewidth}
    \centering
    \vspace{-10pt}
    \includegraphics[width=\linewidth]{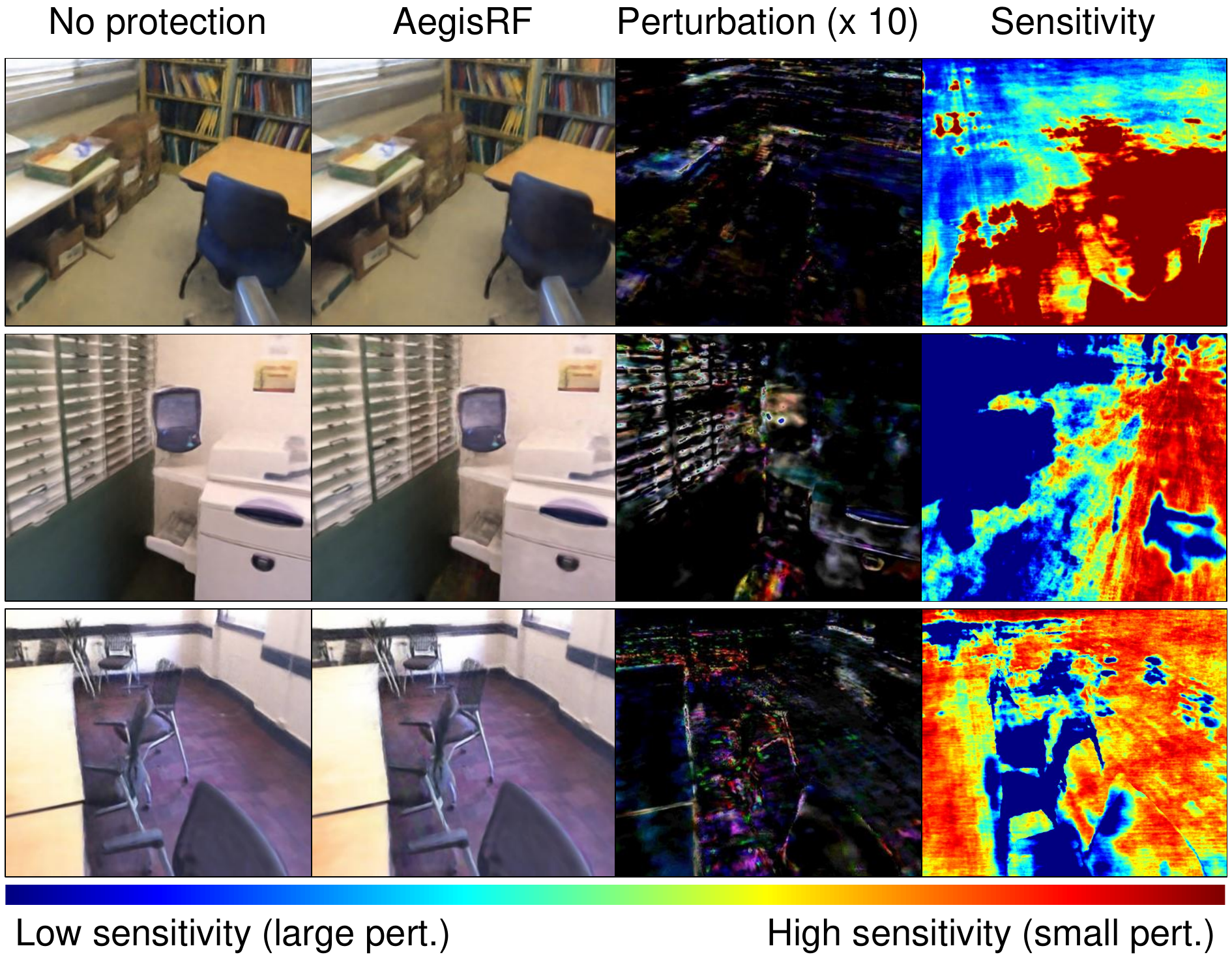}
    \vspace{-20pt}
    \caption{
    Images rendered from NeRF (col. 1) and \sys (col. 2), pixel-wise perturbation (col. 3), and sensitivity averaged along each ray of the pixel (col. 4).
    }
    \vspace{-5pt}
    \label{fig:pert-eps-viz}
\end{wrapfigure}

\vspace{0.5ex}\noindent
\textbf{Visualization of sensitivity.}
In Fig.~\ref{fig:pert-eps-viz}, we visualize the images rendered from the original NeRF (col. 1) and \sys (col. 2) along with the pixel-wise perturbations (col. 3).
The perturbations show that our \sys tends to affect the edges of more complex surfaces, which is consistent with prior findings that distortions on high-frequency textures are generally less perceptible than those on simpler regions~\cite{legge, vq-assessment, thicket}.
We also visualize the sensitivity predicted by Sensitivity Field by averaging values on sampled points along each ray (col. 4).
Sensitivity Field tends to learn lower sensitivity on regions containing objects with complex textures (\eg, chairs), suggesting an implicit alignment with observations from previous studies~\cite{legge, vq-assessment, thicket}, though it is not explicitly designed for this behavior.
It also predicts higher sensitivity along rays cast through empty spaces, thus avoiding perturbations on low-density areas where perturbations can introduce new visible artifacts.

In supplementary materials, we provide ablation studies (Sec.~\ref{sec:ablation}). We also provide additional analysis including the comparison of learned sensitivity with different variations, the effects of soft clamping strategy, and the effects of density perturbation (Sec.~\ref{sec:addtional-analysis}). 

\section{Conclusion}
In this work, we introduce \sys, an adversarial framework to protect the intellectual property of NeRFs by undermining the performance of target unauthorized downstream models.
We design the Perturbation Field, which injects adversarial perturbations to the pre-rendering color and density outputs of NeRFs, thus allowing our approach to effectively safeguard their IP from a wide range of downstream tasks and modalities. 
We also introduce a novel Sensitivity Field that adaptively constrains the magnitudes of geometric perturbations based on their impact on rendering quality to preserve the visual fidelity.
Through comprehensive experimental evaluations, we verify the ability of \sys to protect NeRFs from a variety of downstream applications, including multi-view image classification and voxel-based 3D localization, without compromising their rendering quality.
We hope that our work contributes towards more secure deployment of NeRFs as a 3D data representation.

\vspace{2ex}\noindent
\textbf{Acknowledgments}
This work was supported by Institute of Information \& communications Technology Planning \& Evaluation (IITP) grant funded by the Korea government(MSIT) (No.RS-2025-25443318, Physically-grounded Intelligence: A Dual Competency Approach to Embodied AGI through Constructing and Reasoning in the Real World and RS-2023-00237965, Recognition, Action and Interaction Algorithms for Open-world Robot Service) and the National Research Foundation of Korea(NRF) grant funded by the Korea government(MSIT) (No. RS-2023-00208506).

\clearpage
\newcommand{\appendixnumbering}{
    \renewcommand{\thefigure}{S\arabic{figure}}
    \renewcommand{\thetable}{S\arabic{table}}
}

\appendix
\appendixnumbering

\clearpage
\setcounter{page}{1}
\setcounter{table}{0}
\setcounter{figure}{0}

\begin{center}
    {\LARGE \bfseries \textit{Supplementary Materials}: \\
    AegisRF: Adversarial Perturbations Guided with Sensitivity for Protecting Intellectual Property of Neural Radiance Fields}\\[1em]
\end{center}

In this supplementary material, we provide additional details and experiment results not included in the main paper.
In Sec.~\ref{sec:additional-details}, we include additional implementation details on our method and experimental setups.
In Sec.~\ref{sec:transferability}, we analyze the ability of our method to protect radiance fields from target tasks with unknown perception models, \ie, cross-model transferability.
In Sec.~\ref{sec:common-transformations}, we evaluate the robustness of our method against common transformations that could be employed by the adversary to break the adversarial effects of our perturbations.
In Sec.~\ref{sec:ablation}, we perform ablation studies on the two components of our \sys: Perturbation and Sensitivity Fields.
In Sec.~\ref{sec:addtional-analysis}, we provide additional analysis on our method, including the effects of learned sensitivity, computational costs, the effects of our soft clamping strategy (Eq.~\ref{eq:soft-clamp}), and the effects of density perturbation.

\section{Implementation Details}
\label{sec:additional-details}
\vspace{0.5ex}\noindent
\textbf{Model architecture.}
First, we explain the architectures of our Perturbation Field $G_{p}$ and Sensitivity Field $G_{s}$ in more detail.
For $G_{p}$, we apply position encoding of dimension $12$ to position $\mathbf{x}$, which is passed into $4$ fully-connected ReLU layers, each with $256$ channels.
This output is further passed through an additional linear layer to output the density perturbation $\delta^\sigma$ and a different linear layer of channel $256$ to output a feature vector.
We then apply position encoding of dimension $4$ to viewing direction $\mathbf{d}$, which is concatenated with the feature vector and passed through an additional linear layer with channel $128$ with a ReLU activation, followed by a final linear layer to output the color perturbation $\delta^\mathbf{c}$. 
Sensitivity Field $G_{s}$ has a similar architecture with $G_{p}$, except that it only uses the $4$ fully-connected ReLU layers and an additional linear layer to output the sensitivity $s$ given only the positional encoded position.
Since the constraint is applied only on the density perturbation, it is agnostic to the viewing direction of the point.

Following vanilla NeRF~\cite{nerf}, we apply a positional encoding to the input position and direction of Perturbation and Sensitivity Fields.
Also following the well-known approach in NeRFs, we additionally use a per-image embedding vector that encodes the identity of each camera used to capture the image~\cite{depth-nerf, nerf-w}.
This camera embedding is concatenated with the positional encoded viewing direction, which is passed as an input to $G_{p}$.

\vspace{0.5ex}\noindent
\textbf{Implementation details.}
We train \sys for $30$k iterations using an Adam optimizer with \texttt{lr} = \texttt{5e-4}, $\beta_1 = 0.9$, $\beta_2 = 0.999$ and Cosine Annealing scheduler~\cite{cosine-annealing}.
We also train Adv-FT, which adversarially fine-tunes NeRF, for $30$k iterations and use the learning rates set by original NeRF methods~\cite{nerf, depth-nerf}.

\vspace{0.5ex}\noindent
\textbf{Dataset details.}
We used $8$ scenes from Realistic Synthetic objects from NeRF~\cite{nerf} with image size $224 \times 224$ and $10$ scenes from ScanNet dataset~\cite{scannet} with image size $624 \times 468$.

\begin{table*}[t]
    \centering
    \resizebox{0.7\linewidth}{!}{
    \begin{tabular}{c|c|cccc}
    \specialrule{2pt}{\aboverulesep}{\belowrulesep}
    \multicolumn{6}{c}{Multi-view classification} \\
    \specialrule{1pt}{\aboverulesep}{\belowrulesep}
    \multirow{2}{*}{Target model} & \multirow{2}{*}{Surrogate model} & \multicolumn{4}{c}{Disruption efficacy} \\
    \cline{3-6}

    & & \multicolumn{4}{c}{Acc. (\%)} \\
    \specialrule{1pt}{\aboverulesep}{\belowrulesep}
    \multirow{4}{*}{ViT-B} & No protection & \multicolumn{4}{c}{99.00} \\
    \cline{2-6}
    & ViT-B & \multicolumn{4}{c}{1.88} \\
    & Swin-B & \multicolumn{4}{c}{19.63} \\
    & Mixer-B & \multicolumn{4}{c}{7.25} \\
    \midrule
    \multirow{4}{*}{Swin-B} & No protection & \multicolumn{4}{c}{99.88} \\
    \cline{2-6}
    & ViT-B & \multicolumn{4}{c}{41.00} \\
    & Swin-B & \multicolumn{4}{c}{0.00} \\
    & Mixer-B & \multicolumn{4}{c}{6.00} \\
    \midrule
    \multirow{4}{*}{Mixer-B} & No protection & \multicolumn{4}{c}{90.75} \\
    \cline{2-6}
    & ViT-B & \multicolumn{4}{c}{44.75} \\
    & Swin-B & \multicolumn{4}{c}{18.75} \\
    & Mixer-B & \multicolumn{4}{c}{0.50} \\
    \specialrule{2pt}{\aboverulesep}{\belowrulesep}

    \multicolumn{6}{c}{3D localization} \\
    \specialrule{1pt}{\aboverulesep}{\belowrulesep}
    \multirow{2}{*}{Target model} & \multirow{2}{*}{Surrogate model} & \multicolumn{4}{c}{Disruption efficacy} \\
    \cline{3-6}
    & & $\text{Recall}_{25}$ (\%) & $\text{Recall}_{50}$ (\%) & $\text{AP}_{25}$ (\%) & $\text{AP}_{50}$ (\%) \\
    \specialrule{1pt}{\aboverulesep}{\belowrulesep}
    \multirow{4}{*}{Swin-S} & No protection & 95.44 & 66.63 & 59.94 & 44.35 \\
    \cline{2-6}
    & Swin-S & 48.64 & 4.77 & 3.24 & 0.20 \\
    & VGG & 94.33 & 52.66 & 47.01 & 24.67 \\
    & ResNet & 93.49 & 54.73 & 51.09 & 25.83 \\
    \midrule
    \multirow{4}{*}{VGG} & No protection & 86.70 & 47.96 & 45.05 & 24.07 \\
    \cline{2-6}
    & Swin-S & 80.58 & 35.63 & 31.05 & 12.64 \\
    & VGG & 37.16 & 7.78 & 5.08 & 4.03 \\
    & ResNet & 83.30 & 32.14 & 33.89 & 12.86 \\
    \midrule
    \multirow{4}{*}{ResNet} & No protection & 81.09 & 41.60 & 34.12 & 13.43 \\
    \cline{2-6}
    & Swin-S & 74.90 & 21.92 & 19.87 & 3.31 \\
    & VGG & 67.25 & 13.03 & 7.42 & 1.76 \\
    & ResNet & 39.51 & 2.00 & 0.86 & 0.00 \\
    \specialrule{2pt}{\aboverulesep}{\belowrulesep}

    \end{tabular}
    }
    \caption{
    Transferability of \sys from a surrogate perception model to unknown target models in multi-view classification and 3D localization.
    First column represents the target model on which \sys is evaluated, and second column represents the surrogate model used to train \sys.
    }
    \label{tab:transferability}
\end{table*}

\section{Cross-Model Transferability}
\label{sec:transferability}
In this section, we report additional cross-model transferability~\cite{ada} results on 3D localization (NeRF-RPN~\cite{nerf-rpn}).
The results of these evaluations are presented in Table~\ref{tab:transferability}.
While a bit reduced compared to multi-view classification, our \sys also demonstrates transferability across different backbones for 3D localization and 3D segmentation.
For instance, \sys trained to undermine NeRF-RPN with a VGG-based backbone can also degrade the $\text{Recall}_{50}$ of NeRF-RPN with a ResNet-based backbone from $41.60\%$ to $13.03\%$.

These results underscore the practical utility of \sys in protecting radiance fields from unknown perception models in downstream tasks.
However, the current level of transferability achieved by \sys, while promising, remains limited, particularly when applied to more complex tasks such as 3D localization or across perception models with significantly different architectures (\eg, Swin-based and CNN-based backbones in 3D localization).
To enhance its practicality, future research should focus on developing advanced techniques to improve the disruption efficacy of \sys over a wider range of perception models and downstream tasks by incorporating more diverse surrogate models during training or by adopting techniques from existing black-box adversarial attacks.

\begin{figure*}
    \centering
    \includegraphics[width=\linewidth]{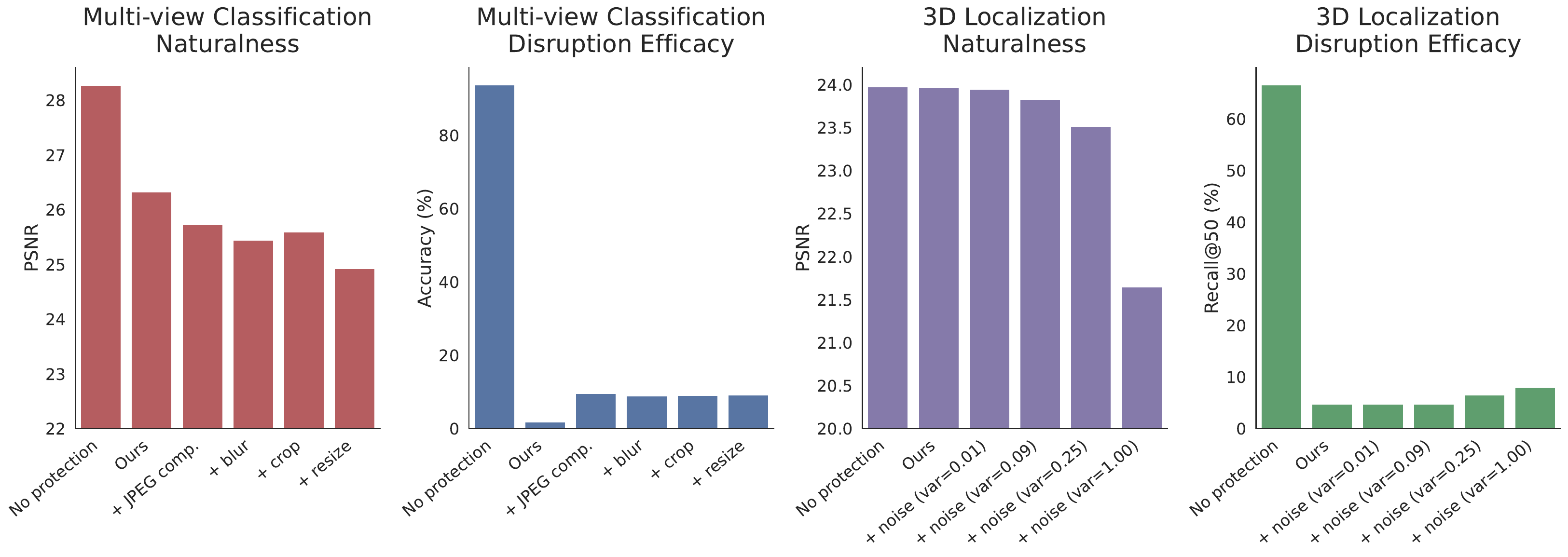}
    \vspace{-10pt}
    \caption{
    Robustness of our \sys against common transformations.
    Our perturbations are robust against common transformations. 
    While these transformations slightly degrade the disruption efficacy on target model, they also degrade the rendering quality, making it difficult to neutralize the IP protection provided by \sys while preserving visual fidelity.
    }
    \label{fig:common-transformations}
\end{figure*}

\section{Robustness to Common Transformations}
\label{sec:common-transformations}
Recent studies have shown that adversarial perturbations against diffusion models can be neutralized using common image transformations, such as JPEG compression or resizing~\cite{common-transformations}.
To assess the real-world applicability of \sys as an IP protection strategy, we evaluate its robustness against these transformations, which adversaries might use to counteract its effects.
For multi-view classification, we apply common transformations including JPEG compression at 50\% quality, Gaussian blurring, cropping 10\% from each of the four margins of an image, and downsampling followed by upsampling by a factor of 2. For 3D localization, we introduce Gaussian noise at varying magnitudes by adjusting the variance.

As shown in Fig.~\ref{fig:common-transformations}, our \sys is robust against common transformations.
For example, it only undergoes $7.12\%$p accuracy increase against blurring in multi-view classification and $3.26\%$p Recall@50 increase against Gaussian noise with variance $= 1$.
These transformations also degrade the rendering quality, showing that it is difficult to neutralize our \sys while preserving the visual fidelity with simple transformations.

\section{Ablation Studies}
\label{sec:ablation}
\outline{0. Introduction on subsection
}
We perform ablation studies on the two components of our \sys: Perturbation and Sensitivity Fields.

\begin{table}[t]
    \centering
    \resizebox{0.7\columnwidth}{!}{
    \begin{tabular}{c|ccc|cccc}
    \specialrule{2pt}{\aboverulesep}{\belowrulesep}
    \multirow{2}{*}{Method} & \multicolumn{3}{c|}{Naturalness} & \multicolumn{4}{c}{Disruption efficacy} \\
    \cline{2-8}
    & PSNR & SSIM & LPIPS & $\text{Recall}_{25}$ & $\text{Recall}_{50}$ & $\text{AP}_{25}$ & $\text{AP}_{50}$ \\
    \specialrule{2pt}{\aboverulesep}{\belowrulesep}
    No protection & 23.98 & 0.7884 & 0.3186 & 95.44 & 66.63 & 59.94 & 44.35 \\
    \hline
    pert. on $\mathbf{c}$ & 23.98 & 0.7810 & 0.3410 & 89.90 & 59.80 & 51.76 & 37.50 \\
    pert. on $\sigma$ & 23.94 & 0.7795 & 0.3339 & 88.90 & 58.95 & 49.41 & 34.63 \\
    pert. on $\mathbf{c}$ \& $\sigma$ (Ours) & 23.97 & 0.7687 & 0.3565 & 48.64 & 4.77 & 3.24 & 0.20 \\
    \specialrule{2pt}{\aboverulesep}{\belowrulesep}
    \end{tabular}
    }
    \caption{
    Ablation studies on perturbing appearance $\mathbf{c}$, geometry $\sigma$, or both.
    }
    \label{tab:ablation-pert}
\end{table}
\vspace{0.5ex}\noindent
\textbf{Perturbation Field.}
As shown in Table~\ref{tab:ablation-pert}, perturbing both appearance and geometry (row 4) leads to the most effective disruption, whereas perturbing only appearance (row 2) or geometry (row 3) results in suboptimal performance.
Perturbing either attribute alone has small impact on misleading the target model, which relies on both appearance and geometric cues.
In contrast, our approach of jointly perturbing both attributes successfully misguides the model while causing only a slight degradation in naturalness.

\begin{table}[t]
    \centering
    \resizebox{0.7\columnwidth}{!}{
    \begin{tabular}{c|ccc|cccc}
    \specialrule{2pt}{\aboverulesep}{\belowrulesep}
    \multirow{2}{*}{Method} & \multicolumn{3}{c|}{Naturalness} & \multicolumn{4}{c}{Disruption efficacy} \\
    \cline{2-8}
    & PSNR & SSIM & LPIPS & $\text{Recall}_{25}$ & $\text{Recall}_{50}$ & $\text{AP}_{25}$ & $\text{AP}_{50}$ \\
    \specialrule{2pt}{\aboverulesep}{\belowrulesep}
    No protection & 23.98 & 0.7884 & 0.3186 & 95.44 & 66.63 & 59.94 & 44.35 \\
    \hline
    No const. & 20.82 & 0.6206 & 0.5137 & 51.92 & 6.02 & 3.34 & 0.13 \\
    const. on $\delta^\mathbf{c}$ & 21.50 & 0.6409 & 0.4918 & 49.82 & 3.25 & 3.20 & 0.02 \\
    const. on $\delta^\sigma$ (Ours) & 23.97 & 0.7687 & 0.3565 & 48.64 & 4.77 & 3.24 & 0.20 \\
    const. on $\delta^\mathbf{c}$ \& $\delta^\sigma$ & 23.99 & 0.7765 & 0.3469 & 64.04 & 12.13 & 10.67 & 1.86 \\
    \specialrule{2pt}{\aboverulesep}{\belowrulesep}
    \end{tabular}
    }
    \caption{
    Ablation studies on applying constraints to appearance perturbations $\delta^\mathbf{c}$, geometry perturbations $\delta^\sigma$, or both.
    }
    \label{tab:ablation-const}
\end{table}
\vspace{0.5ex}\noindent
\textbf{Sensitivity Field.}
We study the effects of the Sensitivity Field by selectively applying the sensitivity-based constraints on perturbations. 
As shown in Table~\ref{tab:ablation-const}, not applying any constraint (row 2) significantly degrades the rendering quality, highlighting the need to explicitly constrain perturbations.
Constraining perturbations on geometry (row 4) is more vital for preserving rendering quality than constraining those on appearance (row 3).
This is because the geometry of radiance fields directly determines point visibility, where even slight changes can cause points to appear or disappear, while appearance perturbations primarily affect color variations without disrupting spatial coherence.
Finally, constraining perturbations on both appearance and geometry (row 5) leads to the best rendering quality but with reduced disruption efficacy.

\begin{table}[t]
    \centering
    \resizebox{0.65\columnwidth}{!}{
    \begin{tabular}{c|ccc|cccc}
    \specialrule{2pt}{\aboverulesep}{\belowrulesep}
    \multirow{2}{*}{Method} & \multicolumn{3}{c|}{Naturalness} & \multicolumn{4}{c}{Disruption efficacy} \\
    \cline{2-8}
    & PSNR & SSIM & LPIPS & $\text{Recall}_{25}$ & $\text{Recall}_{50}$ & $\text{AP}_{25}$ & $\text{AP}_{50}$ \\
    \specialrule{2pt}{\aboverulesep}{\belowrulesep}
    No protection & 23.98 & 0.7884 & 0.3186 & 95.44 & 66.63 & 59.94 & 44.35 \\
    \hline
    Complement & 20.97 & 0.6758 & 0.4620 & \textbf{48.23} & 5.36 & 4.02 & 0.26 \\
    Random & 22.25 & 0.6974 & 0.4398 & 48.57 & \textbf{4.36} & 4.57 & 0.21 \\
    Ours & \textbf{23.97} & \textbf{0.7687} & \textbf{0.3565} & 48.64 & 4.77 & \textbf{3.24} & \textbf{0.20} \\
    \specialrule{2pt}{\aboverulesep}{\belowrulesep}
    \end{tabular}
    }
    \caption{
    Analysis on the sensitivity learned by the Sensitivity Field compared to its complement and random values.
    Best results are in \textbf{bold}.
    }
    \label{tab:eps-analysis}
\end{table}

\begin{table}[t]
    \centering
    \resizebox{0.85\columnwidth}{!}{
    \begin{tabular}{c|c|c|c}
    \specialrule{2pt}{\aboverulesep}{\belowrulesep}
    \textbf{Method} & \textbf{Size} & \textbf{Training Time (30k iterations)} & \textbf{Inference Time (1 image)} \\
    \hline
    NeRF & 2.26 MB & 5.54 hr & 4.46 sec \\
    + \sys (Ours)     & 4.27 MB & 5.81 hr & 6.80 sec \\
    \specialrule{2pt}{\aboverulesep}{\belowrulesep}
    \end{tabular}
    }
    \caption{Comparison of computational costs between original NeRF and our \sys.}
    \label{tab:computation-cost}
\end{table}

\section{Additional Analysis}
\label{sec:addtional-analysis}
In this section, we provide additional analysis on our \sys not covered in the main paper.

\vspace{0.5ex}\noindent
\textbf{Effects of learned sensitivity.}
We compare our method with a ``complement" approach, in which we replace the sensitivity value $s$ with its complement $1 - s$, and the ``random" approach, in which we replace $s$ with a value randomly sampled from a uniform distribution $\mathcal{U}(0, 1)$.
As shown in Table~\ref{tab:eps-analysis}, we can observe that both complement and random approaches lead to significantly degraded naturalness, verifying that the Sensitivity Field learns sensitivity according to the impact of geometric perturbations on rendering quality degradation.

\vspace{0.5ex}\noindent
\textbf{Computational costs.}
In Table~\ref{tab:computation-cost}, we evaluate the computational costs of our approach compared to the original NeRF.
We can observe that our \sys introduces marginal increases in computational costs compared to NeRF.
Because \sys brings in additional Perturbation and Sensitivity Fields along with the original NeRF model, it slightly increases the model size, training time, and inference time.
However, these additional computational costs are justified by the significant improvement in disruption efficacy shown in earlier results (Table~\ref{tab:main-table} of main paper).
The balance underscores the practicality of our \sys, effectively protecting the intellectual property of radiance fields with small computational overheads.

\begin{table*}[t]
    \centering
    \resizebox{0.7\linewidth}{!}{
    \begin{tabular}{c|ccc|cccc}
    \specialrule{2pt}{\aboverulesep}{\belowrulesep}
    \multirow{2}{*}{Method} & \multicolumn{3}{c|}{Naturalness} & \multicolumn{4}{c}{Disruption efficacy} \\
    \cline{2-8}
    & PSNR & SSIM & LPIPS & $\text{Recall}_{25}$ & $\text{Recall}_{50}$ & $\text{AP}_{25}$ & $\text{AP}_{50}$ \\
    \specialrule{2pt}{\aboverulesep}{\belowrulesep}
    No protection & 23.98 & 0.7884 & 0.3186 & 95.44 & 66.63 & 59.94 & 44.35 \\
    \hline
    Hard clip & 23.91 & 0.7679 & 0.3592 & 51.83 & 9.03 & 8.43 & 2.36 \\
    Soft clamp (Ours) & \textbf{23.97} & \textbf{0.7687} & \textbf{0.3565} & \textbf{48.64} & \textbf{4.77} & \textbf{3.24} & \textbf{0.20} \\
    \specialrule{2pt}{\aboverulesep}{\belowrulesep}
    \end{tabular}
    }
    \caption{
    Comparison of our soft clamping strategy and the hard clipping strategy. Best results are in \textbf{bold}.
    }
    \label{tab:soft-clamp}
\end{table*}

\begin{table*}[t!]
    \centering
    \resizebox{0.7\linewidth}{!}{
    \begin{tabular}{c|ccc|cccc}
    \specialrule{2pt}{\aboverulesep}{\belowrulesep}
    \multirow{2}{*}{Method} & \multicolumn{3}{c|}{Naturalness} & \multicolumn{4}{c}{Disruption efficacy} \\
    \cline{2-8}
    & PSNR & SSIM & LPIPS & $\text{Recall}_{25}$ & $\text{Recall}_{50}$ & $\text{AP}_{25}$ & $\text{AP}_{50}$ \\
    \specialrule{2pt}{\aboverulesep}{\belowrulesep}
    No protection & 23.98 & 0.7884 & 0.3186 & 95.44 & 66.63 & 59.94 & 44.35 \\
    \hline
    $\mathbf{c}'$ ($\lambda_\text{nat} = 5$) & 23.29 & 0.7206 & 0.4241 & 53.78 & 10.42 & 9.25 & 7.54 \\
    $\mathbf{c}'$ ($\lambda_\text{nat} = 10$) & 23.75 & 0.7538 & 0.3889 & 65.06 & 18.57 & 16.77 & 6.67 \\
    $\mathbf{c}'$ ($\lambda_\text{nat} = 50$) & \textbf{23.97} & \textbf{0.7810} & \textbf{0.3410} & 89.90 & 59.80 & 51.76 & 37.50 \\
    \hline
    $\mathbf{c}'$ and $\sigma'$ (Ours) & \textbf{23.97} & 0.7687 & 0.3565 & \textbf{48.64} & \textbf{4.77} & \textbf{3.24} & \textbf{0.20} \\
    \specialrule{2pt}{\aboverulesep}{\belowrulesep}
    \end{tabular}
    }
    \caption{
    Analysis on our method of perturbing both color and density compared to perturbing color only with varying weights $\lambda_\text{nat}$ on the naturalness loss $\mathcal{L}_\text{nat}$. Best results are in \textbf{bold}.
    }
    \label{tab:color-pert-only}
\end{table*}
\vspace{0.5ex}\noindent
\textbf{Effects of soft clamping.}
In order to study the effectiveness of our soft clamping strategy (Eq.~\ref{eq:soft-clamp}), we compare our approach with a hard clipping approach\footnote{\texttt{torch.clamp}} where we clip off the density perturbation $\delta^\sigma$ outside the range $[ -(1-s) \cdot \bar{\sigma}, (1-s) \cdot \bar{\sigma} ]$ set by the predicted sensitivity $s$ and the mean density value $\bar{\sigma}$ for all points uniformly sampled from a 3D grid that covers the entire scene volume.
As shown in Table~\ref{tab:soft-clamp}, hard clipping leads to degraded performance, especially in terms of disruption efficiency, leading to $5.21\%$p lower $\text{AP}_{25}$ compared to our soft clamping strategy.
This is because when the perturbation occasionally becomes too large, clipping off the perturbation outside the range $[ -(1-s) \cdot \bar{\sigma}, (1-s) \cdot \bar{\sigma} ]$ will prevent the gradient flow through the Perturbation Field, hindering its training process and thus leading to suboptimal disruption efficacy.

\vspace{0.5ex}\noindent
\textbf{Effects of density perturbation.}
To further emphasize the significance of density perturbation in radiance field-based downstream tasks, as demonstrated in Sec.~\ref{sec:ablation} of the main paper, we evaluate our approach when applying perturbations only to the color outputs of NeRFs.
In Table~\ref{tab:color-pert-only}, we report the naturalness and disruption efficacy for color-only perturbation ($\mathbf{c}'$) with varying $\lambda_\text{nat}$, which controls the weight of the naturalness loss $\mathcal{L}_\text{nat}$.
We can observe that as $\lambda_\text{nat}$ decreases, the disruption efficacy generally improves, indicating that lower weights on the naturalness loss $\mathcal{L}_\text{nat}$ produce perturbations that better protect the radiance field from downstream tasks. 
For instance, when $\lambda_\text{nat} = 5$, the disruption efficacy reaches its best values (\eg, $9.25$ $\text{AP}_{25}$) compared to other values of $\lambda_\text{nat}$.
However, this improvement comes at the expense of significantly degraded naturalness (\eg, $0.4241$ LPIPS), reflecting a trade-off between maintaining naturalness and achieving effective protection.

Moreover, even with this compromise in naturalness, perturbing only the color fails to surpass our approach, which perturbs both the color and density ($\mathbf{c}'$ and $\sigma'$).
Our method achieves a balanced performance, maintaining a comparable level of naturalness to color-only perturbations while significantly outperforming them in terms of disruption efficacy over all configurations of $\lambda_\text{nat}$.
These findings underscore the critical role of density perturbation in protecting the radiance fields while maintaining acceptable naturalness.

\end{document}